\documentclass[runningheads,a4paper]{llncs}
\pdfoutput=1
\usepackage{makeidx}  
\usepackage{graphicx} 
\usepackage{subfigure}

\usepackage{url}
\urldef{\mailsa}\path|josnar@ece.ubc.ca, hamarneh@cs.sfu.ca, rafeef@ece.ubc.ca|
\newcommand{\keywords}[1]{\par\addvspace\baselineskip
\noindent\keywordname\enspace\ignorespaces#1}
\begin{document}
\frontmatter          
\pagestyle{headings}  

\mainmatter           
\title{Automatic Spatially-Adaptive Balancing of Energy Terms for Image Segmentation}
\titlerunning{Automatic Spatially-Adaptive Balancing of Energy Terms}  
%
\author{Josna Rao$^{1}$, Ghassan Hamarneh$^{2}$, and Rafeef Abugharbieh$^{1}$} 
\authorrunning{J. Rao, G. Hamarneh, and R. Abugharbieh}  
\institute{$^{1}$Biomedical Signal and Image Computing Lab,\\
University of British Columbia, Canada\\
$^{2}$Medical Image Analysis Lab, Simon Fraser University, Canada\\
\mailsa} 

\maketitle 

\begin{abstract}
Image segmentation techniques are predominately based on parameter-laden optimization. The objective function typically involves weights for balancing competing image fidelity and segmentation regularization cost terms. Setting these weights suitably has been a painstaking, empirical process. Even if such ideal weights are found for a novel image, most current approaches fix the weight across the whole image domain, ignoring the spatially-varying properties of object shape and image appearance. We propose a novel technique that autonomously balances these terms in a spatially-adaptive manner through the incorporation of image reliability in a graph-based segmentation framework. We validate on synthetic data achieving a reduction in mean error of 47\% (p-value $<<$ 0.05) when compared to the best fixed parameter segmentation. We also present results on medical images (including segmentations of the corpus callosum and brain tissue in MRI data) and on natural images.
\end{abstract}

\section{Introduction}
Robust automated image segmentation is a highly desirable goal that continues to defy solution. In medical images for example, natural and pathological variability may result in complicated and unpredictable image and shape features. Current segmentation methods are predominantly based on optimization procedures that produce so called `optimal' segmentations at their minimum. Optimization methods typically incorporate a tradeoff between two classes of cost terms: data fidelity and regularization. This actually is the case not only in segmentation, but also in image registration, shape matching, and other computer vision tasks. This basic tradeoff scheme is ubiquitous, relating to Occam's razor and Akaike/Bayesian information criteria \cite{burnham}, and is seen in many forms, such as likelihood versus prior in Bayesian methods \cite{aksel} and loss versus penalty in machine learning \cite{zhao}. Therefore, any advancement in controlling the balance between competing cost terms will benefit many related applications and algorithmic formulations in medical image analysis. Optimization-based segmentation methods that are fragile and highly sensitive to this tradeoff are plentiful, including active contours techniques \cite{kass}\cite{caselles}\cite{osher}\cite{pluemp}, graph cut methods \cite{boykov}, and optimal path approaches \cite{barrett}. For simplifying the exposition of the ideas in this paper, we will adopt the simplified but general form of the cost or energy function:
\begin{equation}
  E(S|I,\alpha,\beta) = \alpha E_{int} (S) + \beta E_{ext} (S|I)
  \label{eq:one}
\end{equation}
where $S$ is the segmentation and $I$ is the image. $E_{int}$ is the internal cost term contributing to the regularization of the segmentation, most often by enforcing some smoothness constraints, in order to counteract the effects of imaging artifacts. $E_{ext}$ is the external cost term contributing to the contour's conformity to desired image features, e.g., edges. The weights $\alpha$ and $\beta$ are typically set empirically by the users based on their judgment of how to best balance the requirements for regularization and adherence to image content. In most cases, this is a very difficult task and the parameters may be unintuitive for a typical non-technical end user, e.g. a clinician, who lacks knowledge of the underlying algorithm's inner working. Also the resultant segmentations can vary drastically based on how this balance is set. Avoiding the practice of ad-hoc setting of such weights is the driving motivation for our work here.

To the best of our knowledge, regularization weights have traditionally been determined empirically and are fixed across the image domain (i.e. do not vary spatially). In Pluempitiwiriyawej \emph{et al} \cite{pluemp}, the weights are changed \emph{as the optimization progresses}, albeit in an ad-hoc predetermined manner. McIntosh and Hamarneh \cite{mcintosh} demonstrated that adapting the regularization weights \emph{across a set of images} is necessary in addressing the variability in real clinical image data. However, neither approach varies the weights spatially across the image and hence are not responsive or adaptive to local features within a single image.

Image regions with noise, weak or missing boundaries, and/or occlusions are commonly encountered in real image data.  For example, degradation in medical images can occur due to tissue heterogeneity (``graded decomposition'' \cite{udupa}), patient motion, or imaging artifacts, e.g. echo dropouts in ultrasound or non-uniformity in magnetic resonance. In such cases and in order to increase segmentation robustness and accuracy, \emph{more regularization is needed in less reliable image regions} which suffer from greater deterioration. Although an optimal regularization weight can be found for a single image in a set \cite{mcintosh}, the same weight may not be optimal for all regions of that image. Spatially adapting the regularization weights provide greater control over the segmentation result, allowing it to adapt not only to images with spatially varying noise levels and edge strength, but also to objects with spatially-varying shape characteristics, e.g. smooth in some parts and jagged in others.

Some form of spatially adaptive regularization over a single image appeared in a recent work by Dong \emph{et al} \cite{dong}. For segmenting an aneurysm, they varied the amount of regularization based on the surface curvature of a \emph{pre-segmented} vessel. The results demonstrated improvements due to adaptive regularization. However, the regularization weights did not rely on the properties of the image itself, which limited the generality of the method. Kokkinos \emph{et al} \cite{kokkinos} investigated the use of adaptive weights for the task of separating edge areas from textured regions using a probabilistic framework, where the posterior probabilities of edge, texture, and smoothness cues were used as weights for curve evolution. Similarily, Malik \emph{et al} \cite{malik} and, very recently, Erdem and Tari \cite{erdem} tackled the problem of texture separation and selected weights based on data cues. However, while these methods focused on curve evolution frameworks, our current work focuses on graph-based segmentation. Additionally, we emphasize balancing the cost terms by adapting regularization for images plagued by noisy and weak or diffused edge problems rather than textured patterns in natural images, which we leave as future work.

In this paper, we advocate the strong need for spatially-adaptive balancing of cost terms in an automated, robust, data-driven manner to relax the requirement on the user to painstakingly tweak these parameters. We also demonstrate how existing fixed-weight approaches (even if globally optimized) are often inadequate for achieving accurate segmentation. To address the problem, we propose a novel data-driven method for spatial adaptation of optimization weights. We develop a new spectral flatness measure of local image noise to balance the energy cost terms at every pixel, without any prior knowledge or fine-tuning.

We validate our method on synthetic, medical, and natural images and compare its performance against two alternative approaches for regularization: using the best possible spatially-fixed weight, and using the globally optimal set of spatially-varying weights as found automatically through dynamic programming.

The rest of this paper is organized as follows; Section \ref{methods} presents a brief overview of our segmentation process, the formulation for our proposed reliability method along with a formulation of the globally optimum graph search approach. Section \ref{results} presents qualitative comparisons of our method to both globally-optimal and fixed parameter-based methods and reports quantitative analysis of the resulting error. Section \ref{conclusion} presents our conclusions and an overview of future work planned.
\section{Methods}
\label{methods}
Our formulation employs energy-minimizing boundary-based segmentation, where the objective is to find a contour that correctly separates an object from background. We begin by formulating the energy of a contour and specifying how the regularization term is weighted in our definition. We then present our approach for a data-driven spatially adaptive regularization method, and end the section with a brief discussion of the globally-optimum parameter method. 
\subsection{Energy-Minimizing Segmentation}
\label{segment}
We embed a parametric contour $C(q)= C(x(q),y(q)):[0,1]\rightarrow\Omega\subset\mathrm{\mathbf{R^2}}$ in image $I:\Omega\rightarrow\mathrm{\mathbf{R}}$. We use a single adaptive weight $w(q)\in[0,1]$ that varies over the length of the contour and re-write (\ref{eq:one}) as:
\begin{equation}
  E(C(q),w(q)) = \int_{0}^{1}\left(w(q) E_{int} (C(q)) + (1-w(q)) E_{ext} (C(q))\right) dq
  \label{eq:two}
\end{equation}
where
\begin{equation}
	E_{ext}(C(q)) = 1-\left|\nabla I(C(q))\right|/\max\limits_{\Omega} \left|\nabla I(C(q))\right|
\end{equation}
penalizes weak boundaries and 
\begin{equation}
	E_{int}(C(q)) = \left|dC(q)/dq \right|
\end{equation}
is the length of the contour. To minimize $E$ with respect to $C(q)$ in (\ref{eq:two}), we adopt a discrete formulation of the optimization problem. We model the image as a graph where each pixel is represented by a vertex $v_i$ and edges $e_{ij} = \langle v_i,v_j \rangle$ that capture the pixel's connectedness (e.g. 8-connectedness in 2D images). A local cost
\begin{equation}
	c_{ij} = c(e_{ij}) = w E_{int} \left(v_i,v_j \right) + \left(1-w \right) E_{ext} (v_i)
\end{equation}
is assigned to each edge $e_{ij}$, where $E_{int} \left( v_i,v_j \right)$ is the Euclidean distances between $v_i$ and $v_j$ (\emph{e.g.} 1 for 4-connected neighbors and $\sqrt{2}$ for diagonal neighbors in 2D). The contour that minimizes the total energy $E = \sum\limits_{e_{ij}\in C} c_{ij}$ represents the optimal solution for the segmentation and is found by solving a minimal path problem, similar to \cite{barrett}, using Dijkstra's algorithm \cite{dij}.
\subsection{Automatic Spatially-Adaptive Balancing of Energy Cost Terms}
Our approach for balancing the cost terms is to gauge the levels of signal vs. noise in local image regions. We estimate the edge evidence $G(x,y)$ and noise level $N(x,y)$ in each region of the image and set $w(x,y)$ in (\ref{eq:two}) such that regions with high noise and low boundary evidence (i.e. low reliability) have greater regularization, and vice versa. Hence, $w(x,y)$ is mapped to image reliability $R(x,y)$ as 
\begin{equation}
	w(x,y) = 1 - R(x,y)  
  \label{eq:three}
\end{equation}
where
\begin{equation}
	R(x,y) = \left( 1 - N(x,y) \right) G(x,y).
	\label{eq:threeB}
\end{equation}
Assuming additive white noise, uncorrelated between pixels, we estimate spatially-varying noise levels $N(x,y)$ using local image spectral flatness (SF). SF is a well-known Fourier-domain measure that has been employed in audio signal processing and compression applications \cite{jayant}\cite{taubman}. SF exploits the property that white noise exhibits similar power levels in all spectral bands and results in a flat power spectrum, whereas uncorrupted signals have power concentrated in certain spectral bands and result in a more impulse-like power spectrum. We extend the SF measure to 2D and measure $N(x,y)$ as
\begin{equation}
	N(x,y) = \frac{\exp \left( \frac{1}{4\pi^2} \int_{-\pi}^{\pi} \int_{-\pi}^{\pi} \ln S
	\left(\omega_x,\omega_y \right)d\omega_x d\omega_y \right)}{\frac{1}{4\pi^2} \int_{-\pi}^{\pi} \int_{-\pi}^{\pi} S
	\left(\omega_x,\omega_y \right)d\omega_x d\omega_y}
  \label{eq:four}
\end{equation}
where $S(\omega_x,\omega_y)$ is the 2D power spectrum of the image and $(\omega_x,\omega_y)$ are spatial frequencies. We use $G(x,y) = \max \left( \left|\nabla I_x(x,y)\right|,\left|\nabla I_y(x,y)\right| \right)$, where $\nabla I_x(x,y)$ and $\nabla I_y(x,y)$ represent the $x$ and $y$ components of the image gradient. We chose this measure rather than the standard gradient magnitude for its rotational invariance in the discrete domain.
\subsection{Non-Contextual Globally Optimal Weights}
\label{globally}
A theoretically appealing and intuitive approach for setting the regularization weight is to optimize $E$ in (\ref{eq:two}) for the \emph{weight} $w(q)$ \emph{itself} in addition to optimizing the contour. In our discrete setting, this involves a `three dimensional' graph search that computes the globally optimal, spatially-adaptive regularization weight $w(q)$, in conjunction with the contour's spatial coordinates, i.e. we optimize\footnote[1]{This is similar in spirit to \cite{poon} and \cite{li} where they also optimize for a non-spatial variable: vessel radius or scale, in addition to the spatial coordinates of the segmentation contour.} $\tilde{C}(q)=\left( x(q),y(q),w(q) \right)$. 

In this formulation, each vertex in the original graph is now replaced by $K$ vertices representing the different choices of the weight value at each pixel. In addition, graph edges now connect vertices corresponding to neighboring image pixels for all possible weights. Note that the optimal path $\tilde{C}(q)$ cannot pass through the same $\left(x(q),y(q) \right)$ for different $w$, i.e. only a single weight can be assigned per pixel. Our graph search abides by this simple and logical constraint. The optimal $C(q)$ and $w(q)$ that globally minimize (\ref{eq:two}) are again calculated using dynamic programming but now on this new $(x,y,w)$ graph. 

There are, however, three main drawbacks to this globally optimum (in $(x,y,w)$) method: (i) it does not explicitly encode image reliability, even though regularization is essential in regions with low reliability; (ii) this approach will encourage a bimodal behavior of the regularization weight: 
\begin{equation}
	w\left( q \right) = \left\{ {\begin{array}{*{40}c}
  0 &&& {E_{int} \left( q \right) > E_{ext} \left( q \right)}  \\
  1 &&& {\mathrm{otherwise}}  \\
 \end{array} } \right\},
\end{equation}
and (iii) combining the weight and segmentation optimization into one process reduces the generality of the method. In short, even though optimal with respect to $E$ in (\ref{eq:two}), the solution is incorrect and, as we later demonstrate, inferior to the spatially adaptive balancing of energy cost terms proposed in Sec. \ref{segment}. Furthermore, by combining the weight optimization and contour optimization processes into one, we reduce the generality of the method. Finding globally-optimal weights for other segmentation frameworks would require significant changes to the energy minimization process.
\section{Results and Discussion}
\label{results}
\begin{figure}
     \centering
     \subfigure[Synthetic sinusoidal image $I(x,y)$]{
          \label{fig:synthI}
          \includegraphics[width=0.45\textwidth,height=20mm]{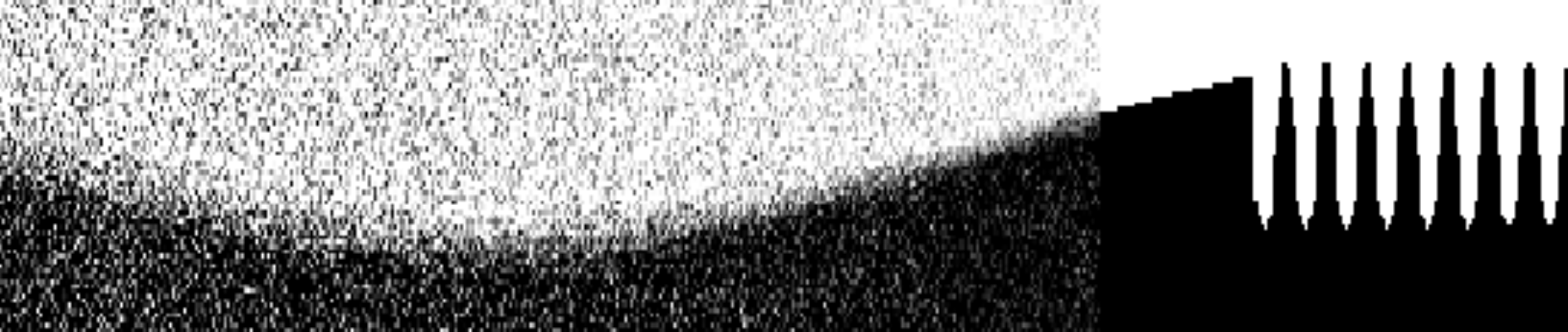}}
     \subfigure[Synthetic sinusoidal image $I(x,y)$]{
          \label{fig:synthI2}
          \includegraphics[width=0.45\textwidth,height=20mm]{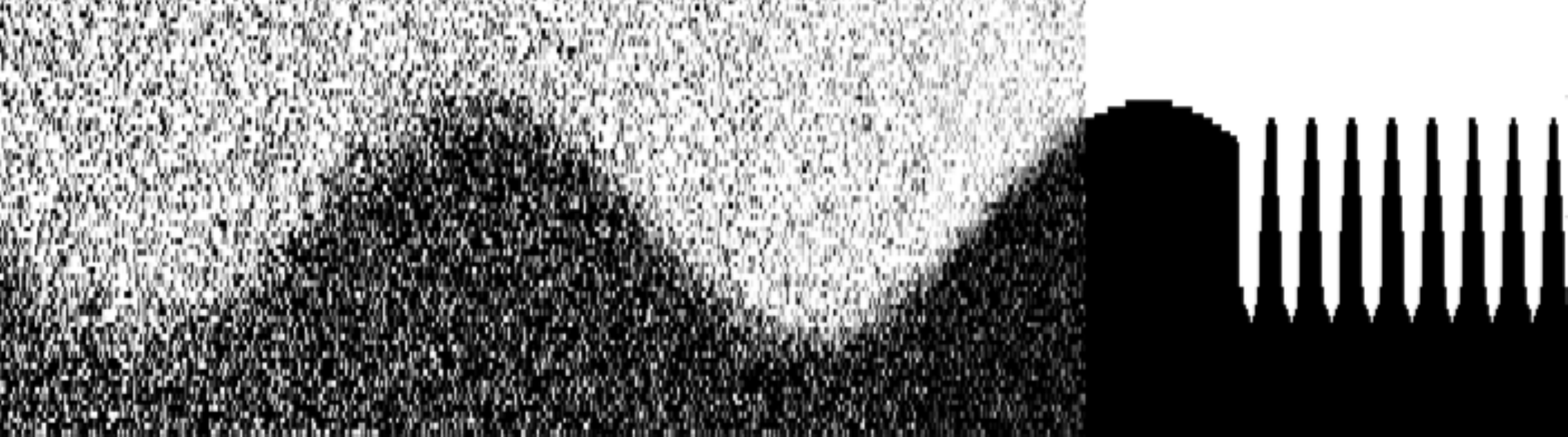}}\\         
     \subfigure[Edge evidence measure $G(x,y)$]{
          \label{fig:synthG}
          \includegraphics[width=0.45\textwidth,height=20mm]{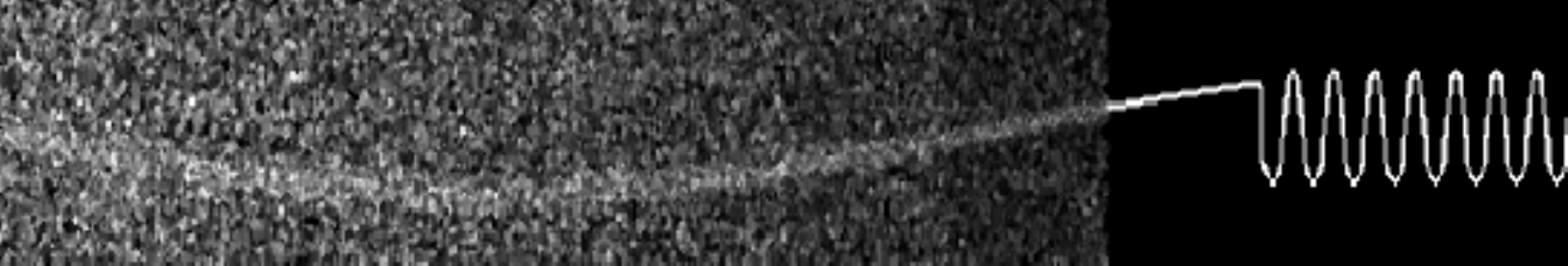}}
     \subfigure[Edge evidence measure $G(x,y)$]{
          \label{fig:synthG2}
          \includegraphics[width=0.45\textwidth,height=20mm]{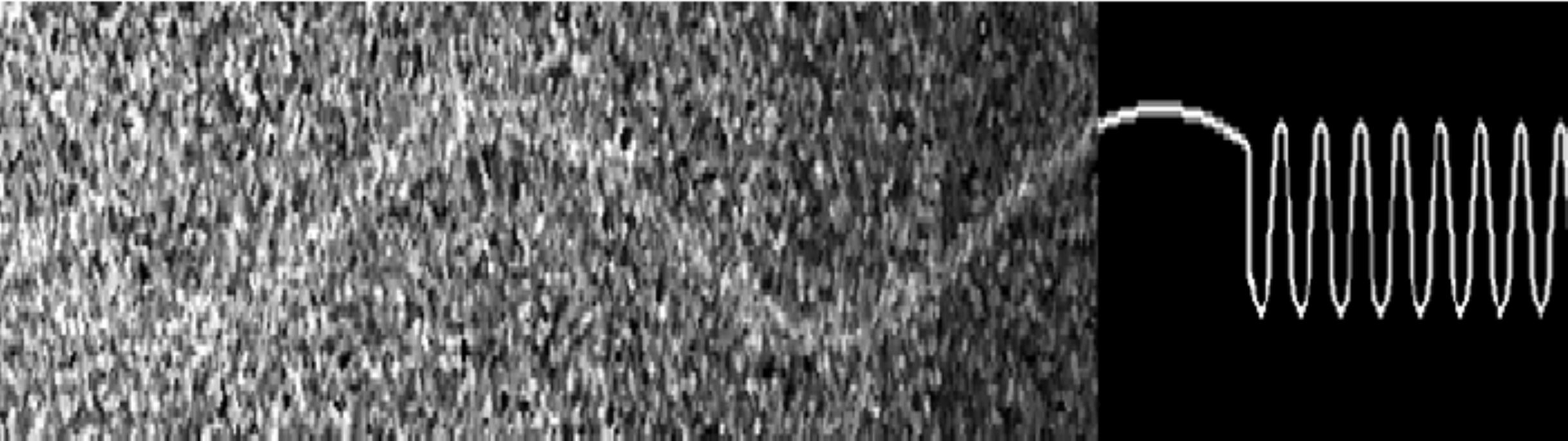}}\\          
     \subfigure[Noise level estimate $N(x,y)$]{
           \label{fig:synthN}
           \includegraphics[width=0.45\textwidth,height=20mm]{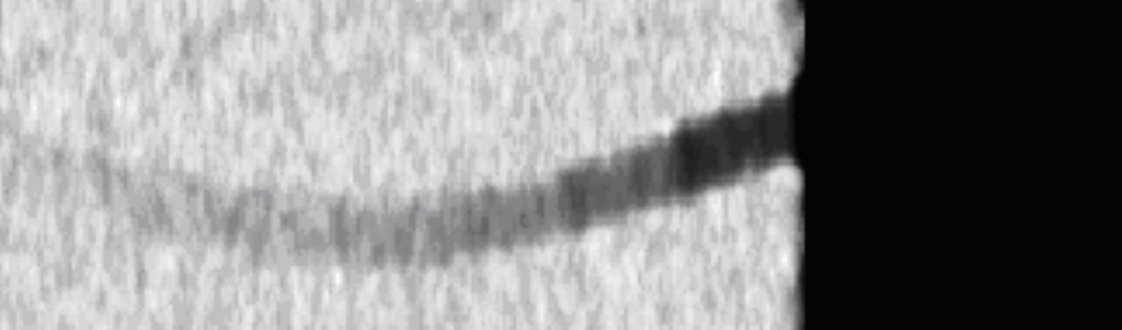}}
     \subfigure[Noise level estimate $N(x,y)$]{
           \label{fig:synthN2}
           \includegraphics[width=0.45\textwidth,height=20mm]{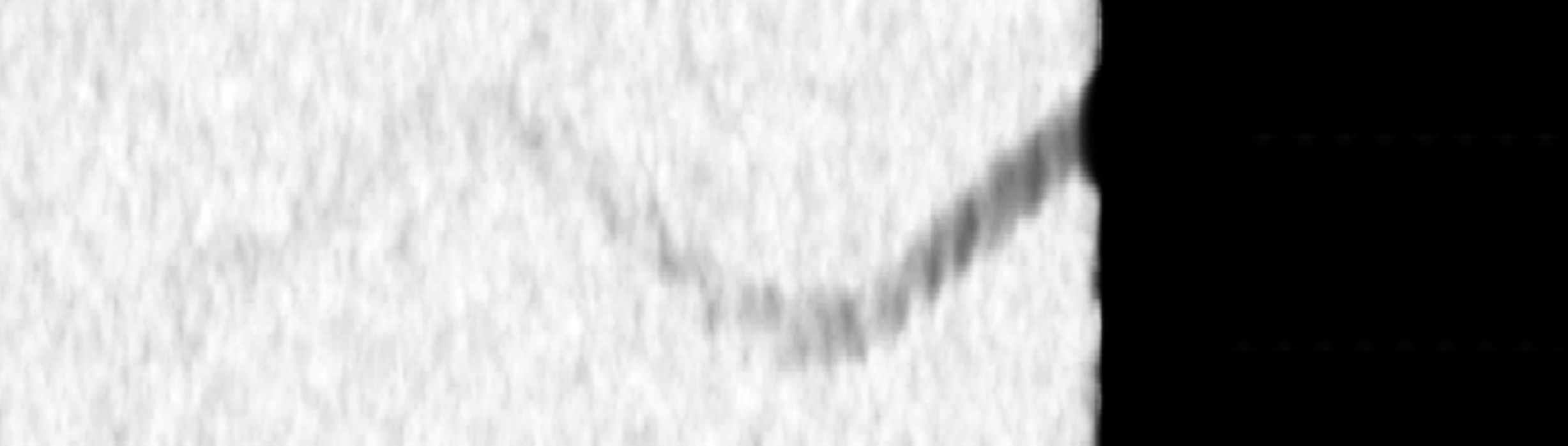}}\\           
     \subfigure[Total reliability measure $R(x,y)$]{
           \label{fig:synthR}
           \includegraphics[width=0.45\textwidth,height=20mm]{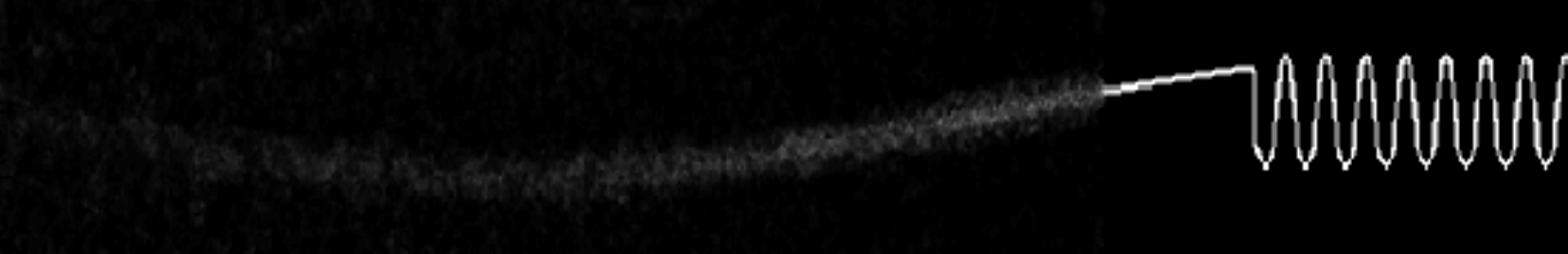}}
     \subfigure[Total reliability measure $R(x,y)$]{
           \label{fig:synthR2}
           \includegraphics[width=0.45\textwidth,height=20mm]{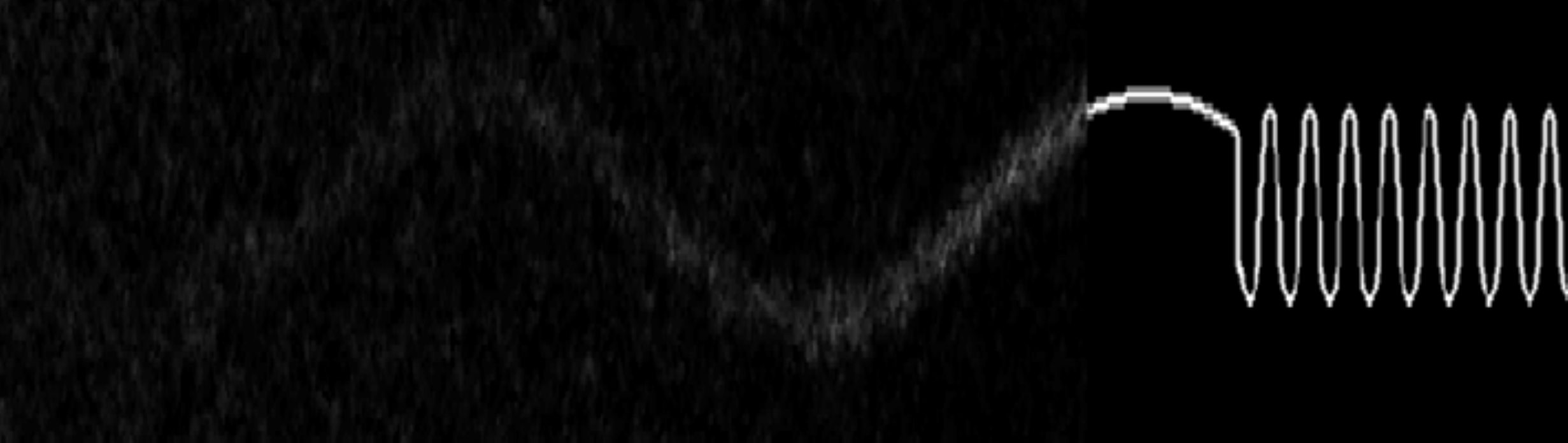}}           
     \caption{Synthetic image with spatially varying noise and blurring (both increasing from right to left) and with changing 	boundary smoothness (smooth on the left and jagged on the right). Black intensities corresponds to 0 and white to 1.}
     \label{fig:synth}
\end{figure}

\begin{figure}
		\centering
	  \subfigure[Segmentation of image in Fig. \ref{fig:synthI}]{
          \label{fig:synthContour1}
          \includegraphics[width=0.95\textwidth]{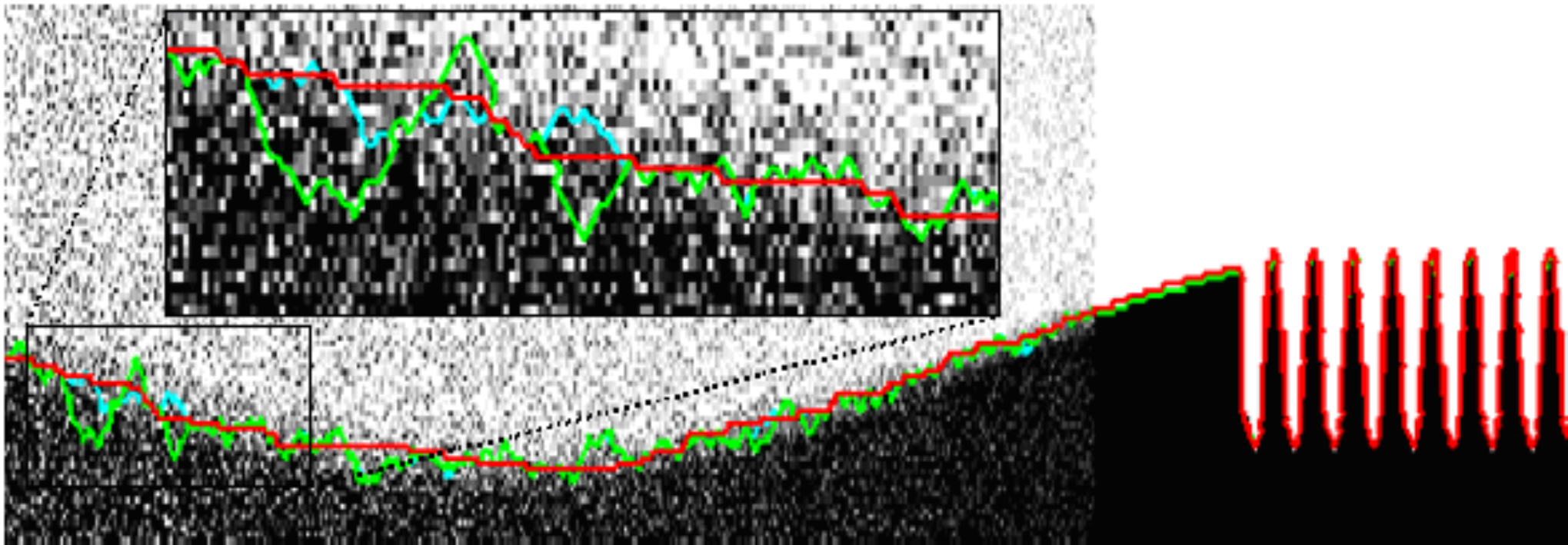}}\\
    \subfigure[Segmentation of image in Fig. \ref{fig:synthI2}]{
          \label{fig:synthContour2}
          \includegraphics[width=0.95\textwidth]{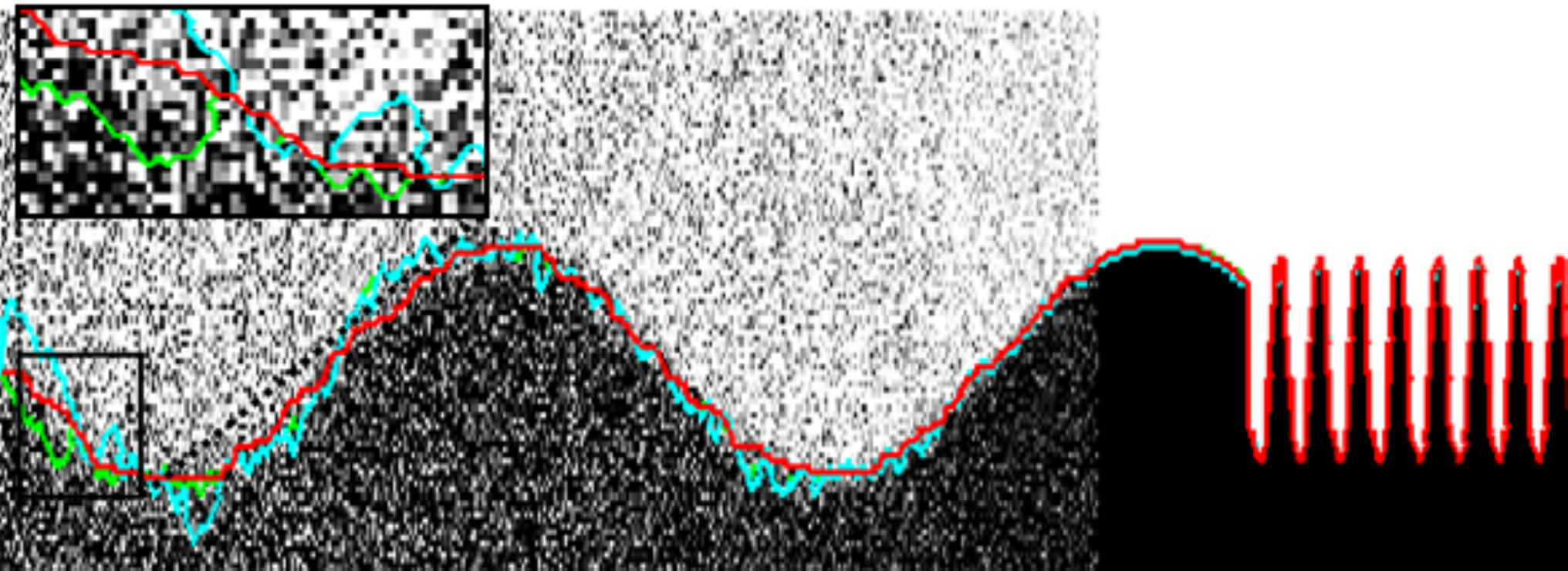}}
		\caption{\emph{Color is essential for proper viewing, please refer to the e-copy}. Contours obtained from: (\emph{red}) proposed adaptive weights, (\emph{green}) lowest-error fixed weight, and (\emph{cyan}) globally optimum weight.}
		\label{fig:synthContour}
\end{figure}

We first performed quantitative tests on 16 synthetic images carefully designed to cover extreme shape and appearance variations (one example is shown in Fig. \ref{fig:synth}). The test data was created by modeling an object boundary as a sinusoidal function with spatially-varying frequency to simulate varying contour smoothness conditions. We then added spatially-varying (non-stationary) additive white Gaussian noise patterns of increasing variance. We also spatially varied the gradient magnitude of the object boundary across each image by applying Gaussian blurring kernels at different scales in different locations. 
Our resulting image reliability measure is exemplified in Fig. \ref{fig:synth} for two synthetic images with the resulting segmentations shown in Fig. \ref{fig:synthContour}. We also show the contour obtained using the globally-optimal weights method (Sec. \ref{globally}), and the contour obtained using a \emph{spatially-fixed} regularization weight, set to the value producing \emph{the smallest} (via brute force search) segmentation error.

We quantitatively examined our method's performance using ANOVA testing on 25 noise realizations of each image in the dataset, where the error was determined by the Hausdorff distance to the ground truth contour. Our method resulted in a mean error (in pixels) of 6.33 (std. dev. 1.36), whereas the best fixed-weight method had a mean error of 12.05 (std. dev. 1.61), and the globally-optimum weight method had a mean error of 33.06 (std. dev. 3.66). Furthermore, for each image, we found our method to be significantly more accurate with all p values $<<$ 0.05.

\begin{figure}
     \centering
     \subfigure[]{
          \label{fig:CC1}
          \resizebox{0.45\textwidth}{40mm}{\includegraphics*[viewport=145 138 766 544]{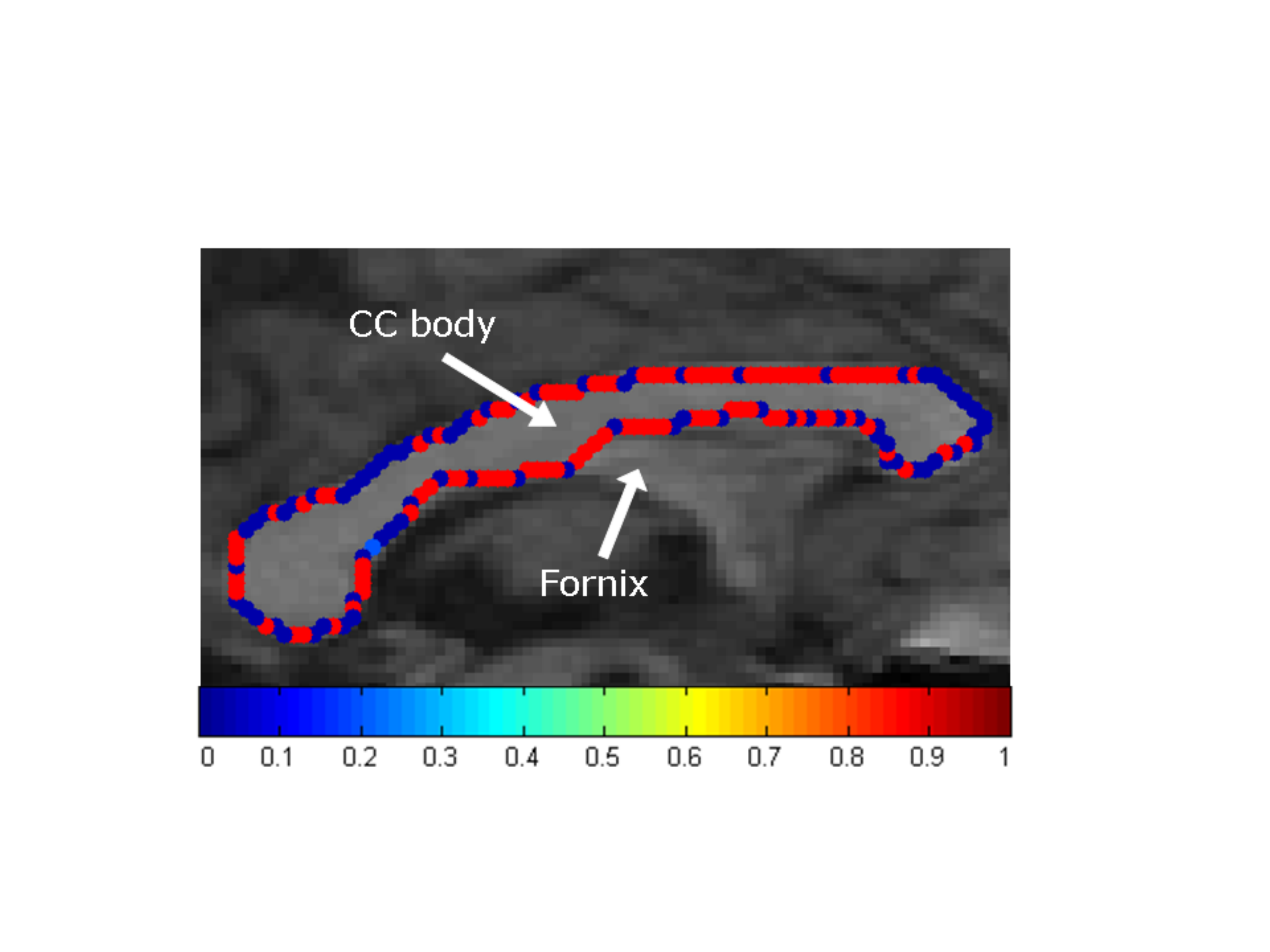}}}
     \subfigure[]{
          \label{fig:CC2}
          \resizebox{0.45\textwidth}{40mm}{\includegraphics*[viewport=145 135 766 544]{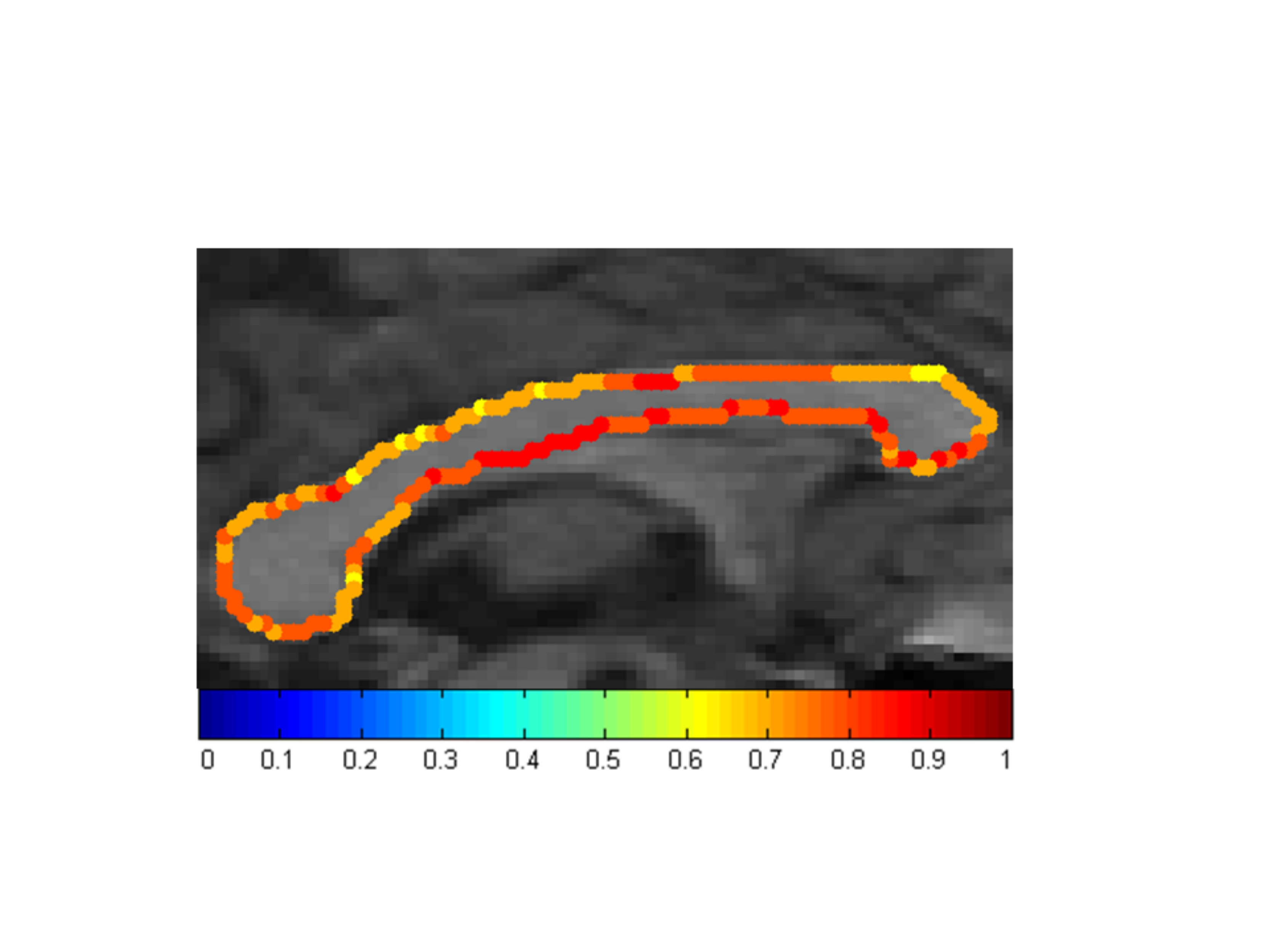}}}
     \caption{(\emph{Color figure, refer to e-copy}). Results of (a) globally-optimum weight method and (b) proposed adaptive-weight method for a corpus callosum MR image. The coloring of the contours reflects the value of the spatially-adaptive weight. The same color map is used for both figures, with pure blue corresponding to $w=0$ for pure red for and $w=1$.}
     \label{fig:CC}
\end{figure}

\begin{figure}
\begin{center}
\resizebox{0.70\textwidth}{50mm}{\includegraphics*[viewport=145 135 766 544]{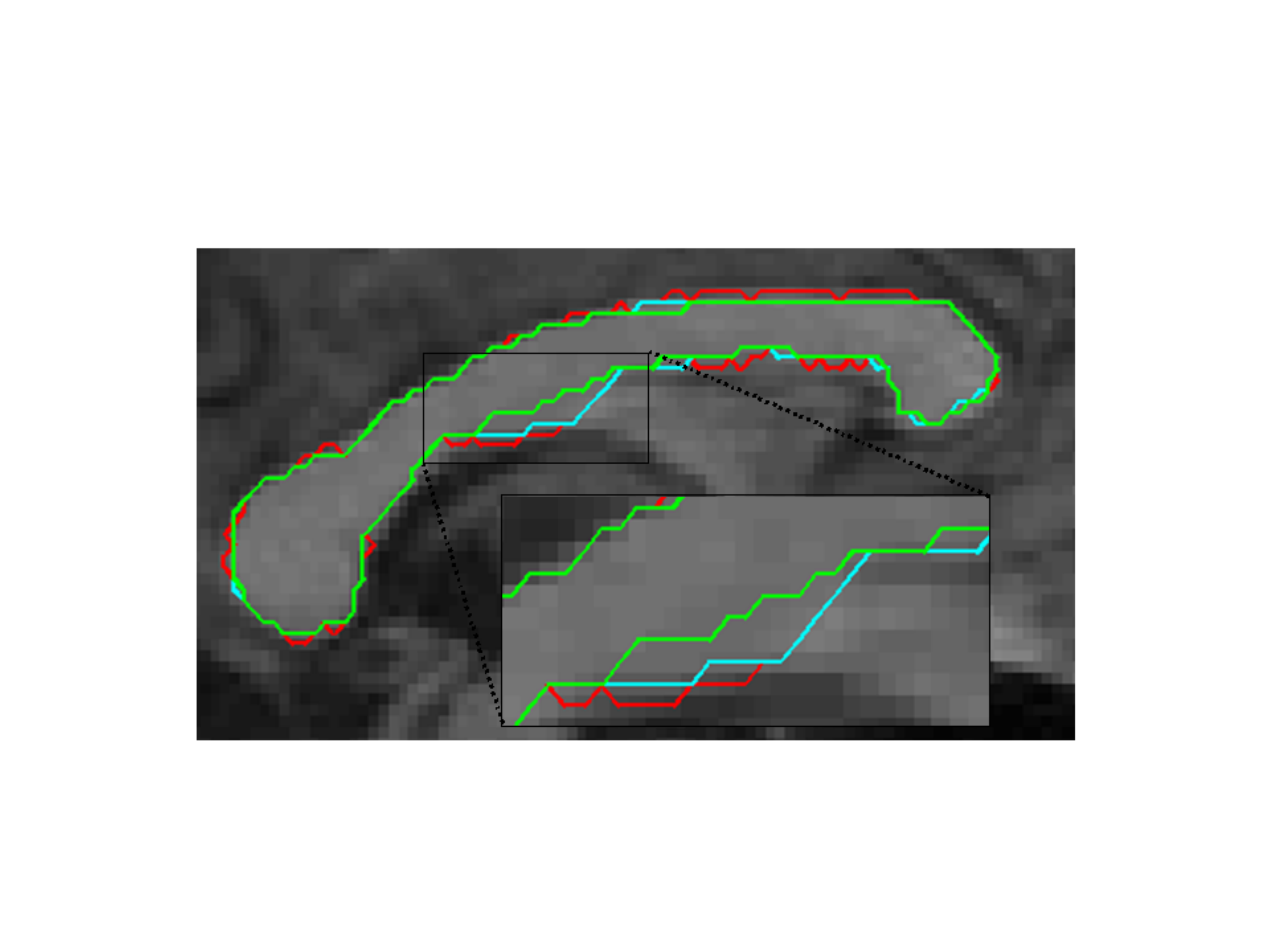}}
\end{center}
\caption{(\emph{Color figure, refer to e-copy}). Segmentation results of the CC in the MR image of Fig. \ref{fig:CC}. \emph{Green} contour is result of the proposed spatially adaptive weight method, \emph{red} contour is result of best fixed-weight method, and \emph{cyan} contour is result of the globally-optimum weight method.}
\label{fig:CCContour}
\end{figure}

We also tested our method on clinical MR images of the corpus callosum (CC), which exhibits the known problem of a weak boundary where the CC meets the fornix (Fig. \ref{fig:CC1}). Note how the contour obtained using globally optimal weights exhibits an optimal, yet undesirable, bimodal behavior (either blue or red in Fig. \ref{fig:CC1}) completely favoring only one of the terms at a time. In comparison, our method automatically boosts up the regularization (stronger red in Fig. \ref{fig:CC2}) at the CC-fornix boundary producing a better delineation. The segmentation results of all three methods for the same image are shown in Fig. \ref{fig:CCContour}.

\begin{figure}
\begin{center}
\scalebox{0.45}{\includegraphics*[viewport=145 135 686 568]{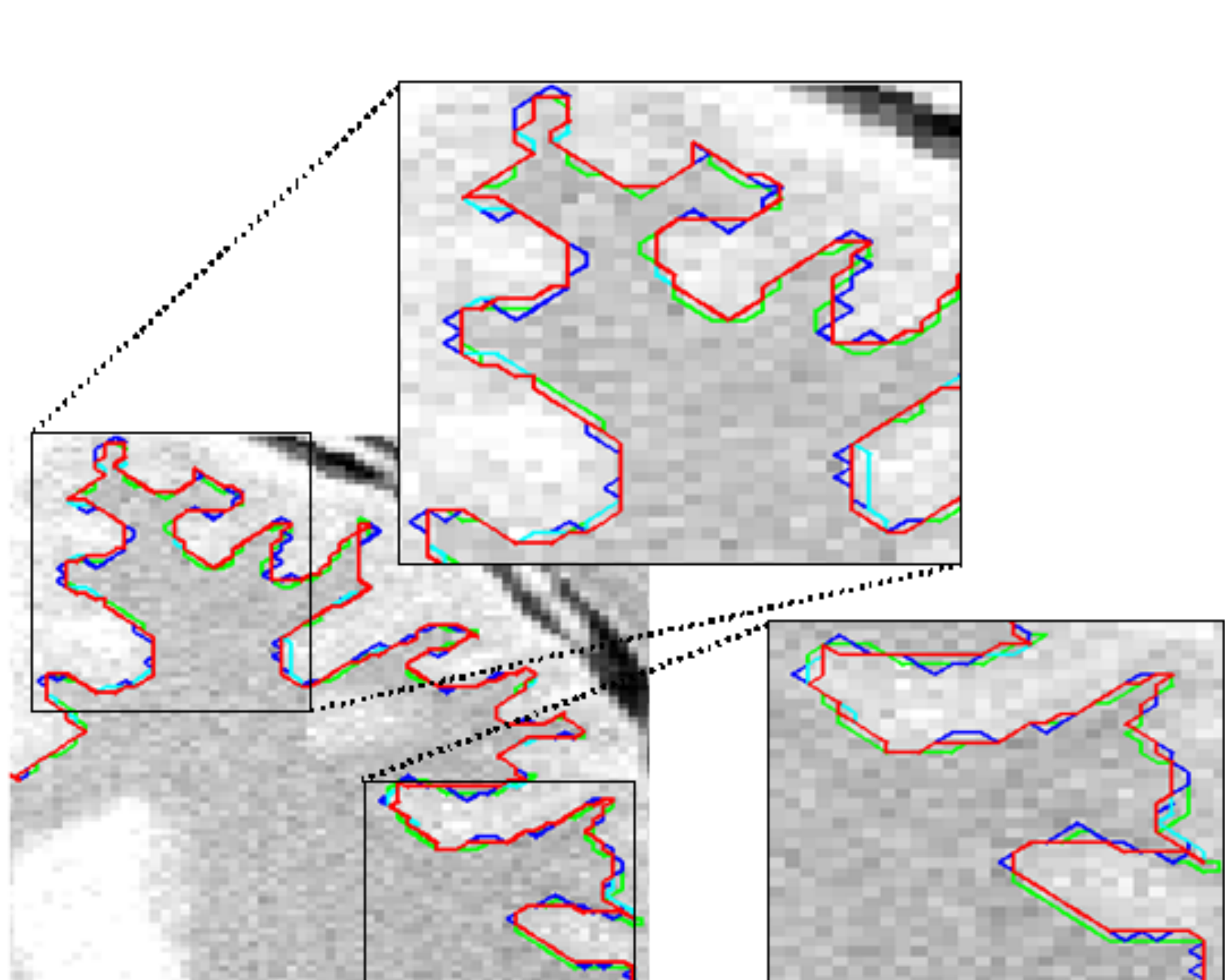}}
\end{center}
\caption{(\emph{Color figure, refer to e-copy}). Contours produced by using proposed adaptive weight (\emph{red}), globally-optimum weight (\emph{cyan}), best fixed-weight (\emph{blue}). Ground truth contour is shown in \emph{green}. Note the improved regularization using our method.}
\label{fig:brain}
\end{figure}

We also tested our method on MR data from BrainWeb \cite{cocosco}. Fig. \ref{fig:brain} shows the segmentation of the cortical surface in a proton density (PD) image with a noise level of 5\%. This example is a difficult scenario due to the high level of noise and low resolution of the image. Our proposed method provided a smoother contour while conforming to the cortical boundary when compared to the other methods (although the difference was not too large).

\begin{figure}
     \centering
     \subfigure[]{
          \label{fig:leaf}
          \resizebox{0.45\textwidth}{60mm}{\includegraphics*[viewport=1 1 231 304]{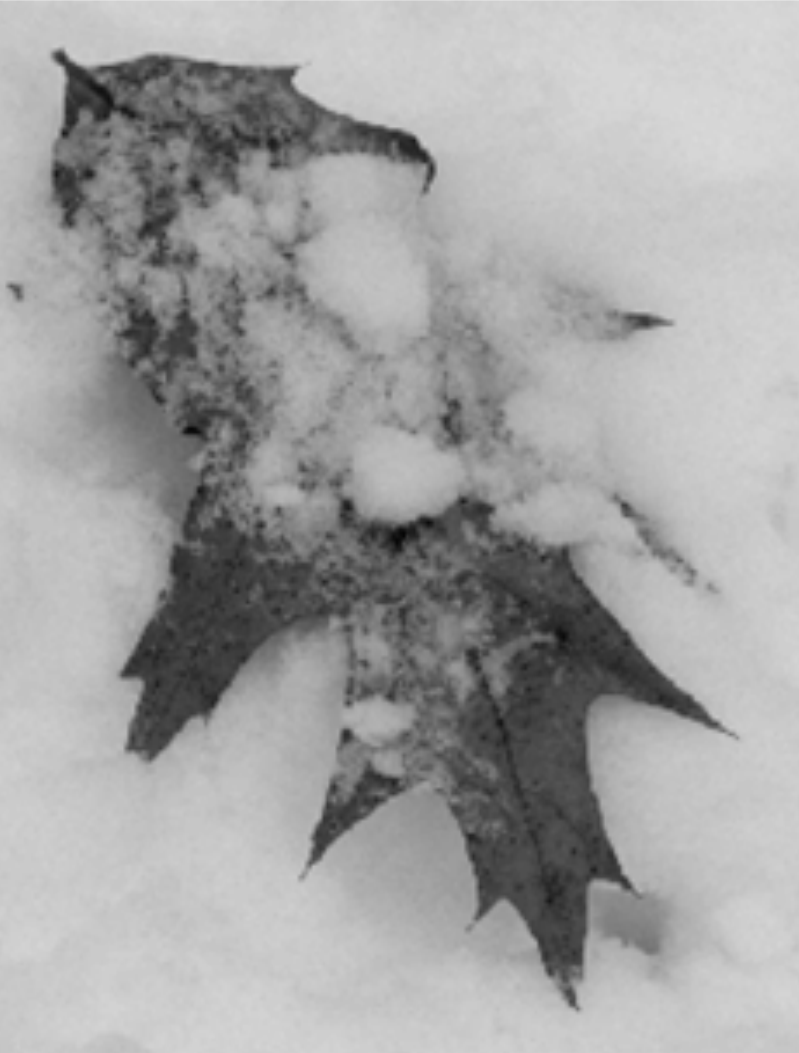}}}
     \subfigure[]{
          \label{fig:leafR}
          \resizebox{0.45\textwidth}{60mm}{\includegraphics*[viewport=1 1 231 304]{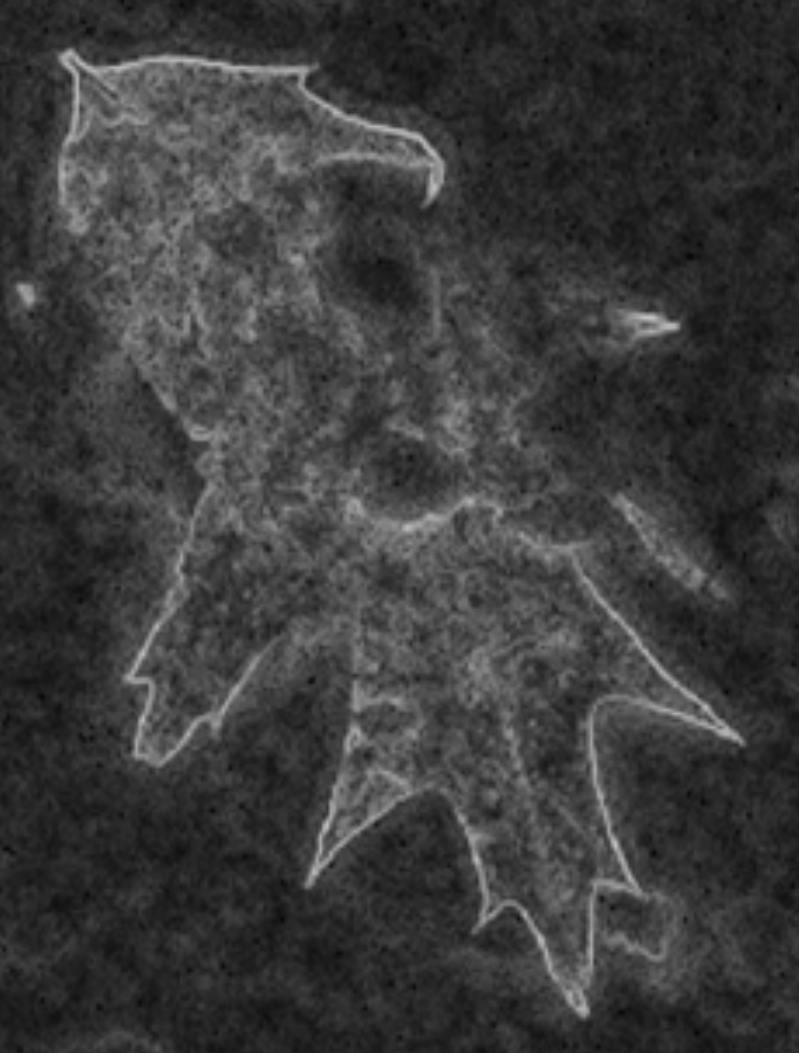}}}
     \caption{Segmenting a natural image. (a) Original leaf image. (b) Reliability calculated by our proposed method.}
     \label{fig:leaves}
\end{figure}

\begin{figure}
\begin{center}
\scalebox{0.45}{\includegraphics*[viewport=1 1 518 422]{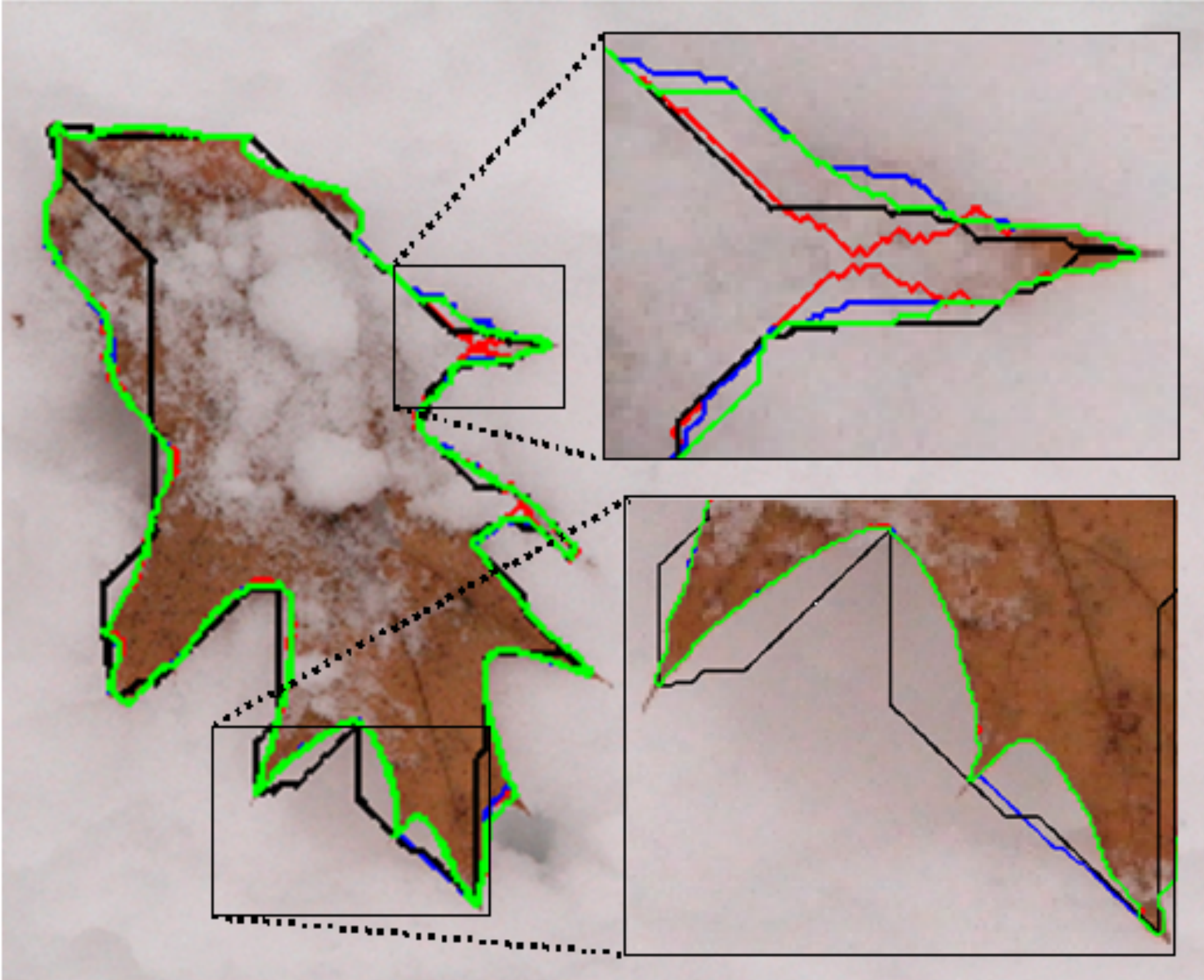}}
\end{center}
\caption{(\emph{Color figure, refer to e-copy}). Segmentation results of the leaf in Fig. \ref{fig:leaf} from our method (\emph{green}), fixed-weight of 1 (\emph{black}), 0.5 (\emph{blue}), and 0 (\emph{red}).}	
\label{fig:leafContour}
\end{figure}

We also tested our method on natural images, such as the leaf shown in Fig. \ref{fig:leaf}. The resulting reliability measure (Fig. \ref{fig:leafR}) has lower image reliability and, hence, higher regularization at the regions of the leaf obscured by snow, whereas reliable boundaries light up (bright white boundary segments). The resulting segmentations are shown in Fig. \ref{fig:leafContour}.

\section{Conclusion}
\label{conclusion}
We proposed a novel approach for addressing a ubiquitous problem that plagues many energy minimization segmentation techniques; how to balance the weights of competing energy terms. Our technique spatially adapts the regularization weight based on local spectral-flatness and data-driven image reliability measures. We emphasize two contributions of our work: (i) regularization must vary spatially and must increase where image evidence is less reliable, and (ii) we discriminate between signal edges (object boundaries) versus noise edges by extending a spectral flatness measure that is well established in audio signal processing to 2D. By analyzing the spectral flatness of the observed signal, a spatially-varying evidence of signal versus noise is derived fully automatically (without any tuning), which we use to spatially adapt the regularization using a clear reliability metric. We note that there is no restriction against using other noise estimation methods in our proposed reliability-modulated regularization, making our method of broader interest. Using a large synthetic dataset exhibiting extreme variations of image deterioration and boundary shape, we demonstrated statistically significant reduction in segmentation error compared to using the best fixed weight, and to a globally-optimal, spatially-varying approach that uses dynamic programming to optimize for the regularization weight in conjunction with contour. 

Our current work focused on minimal-path approaches for segmentation. However, our approach of reliability-based regularization is applicable to a wide range of energy-minimizing segmentation techniques. We are currently extending our approach to other variational and graph-based segmentation approaches such as \cite{kass}\cite{boykov}\cite{chan}. Additionally, we intend to expand our technique to handle energy functionals where multiple weights balance the different energy terms and to explore alternative reliability measures. Another important conclusion from our work is that globally optimal weights do not appear to be desirable. We intend to further explore this issue in more detail in the future.

%
%
\bibliographystyle{splncs}
\bibliography{references}

\end{document}